\documentclass[11pt,a4paper]{article}
\usepackage[hyperref]{acl2019}
\usepackage{times}
\usepackage{latexsym}
\usepackage{graphicx}  %Required
\usepackage{amssymb}
\usepackage{amsmath}
\usepackage{multirow}
\usepackage{bm}
\usepackage[normalem]{ulem}
\usepackage[ruled]{algorithm2e}
\usepackage{natbib}

\usepackage{url}

\aclfinalcopy % Uncomment this line for the final submission
\def\aclpaperid{2762} %  Enter the acl Paper ID here

\title{Reference Network for Neural Machine Translation}

\author{Han Fu\textsuperscript{\rm $\dagger\ddagger$}~~~
	Chenghao Liu\textsuperscript{\rm $\S$}~~~
	Jianling Sun\textsuperscript{\rm $\dagger\ddagger$\thanks{\; Corresponding author: Jianling Sun.}} \\
	\textsuperscript{\rm $\dagger$}Zhejiang University, Hangzhou, China\\
	\textsuperscript{\rm $\ddagger$}Alibaba-Zhejiang University Joint Institute of Frontier Technologies, China\\
	\textsuperscript{\rm $\S$}Singapore Management University, Singapore\\
	\{11821003, sunjl\}@zju.edu.cn\\
	twinsken@gmail.com\\
}

\date{}

\begin{document}
	\maketitle
	\begin{abstract}
		Neural Machine Translation (NMT) has achieved notable success in recent years. Such a framework usually generates translations in isolation. In contrast, human translators often refer to reference data, either rephrasing the intricate sentence fragments with common terms in source language, or just accessing to the golden translation directly. In this paper, we propose a Reference Network to incorporate referring process into translation decoding of NMT. To construct a \emph{reference book}, an intuitive way is to store the detailed translation history with extra memory, which is computationally expensive. Instead, we employ Local Coordinates Coding (LCC) to obtain global context vectors containing monolingual and bilingual contextual information for NMT decoding. Experimental results on Chinese-English and English-German tasks demonstrate that our proposed model is effective in improving the translation quality with lightweight computation cost.
	\end{abstract}
	
	\section{Introduction}
	
	Neural Machine Translation (NMT) has enjoyed impressive success in most large-scale translation tasks \cite{Kalchbrenner2013,Sutskever2014Seq,Cho2014Learn}. Typical NMT model to date is a single end-to-end trained deep neural network that encodes the source sentence into a fixed-length vector and generates the words in the target sentence sequentially. The alignment relationship between source and target sentence is learned by the attention mechanism \cite{bahdanau2015neural,Luong2015Effe}.
	
	Though the framework has achieved significant success, one critical concern is that NMT generates translations in isolation, which leads to translation inconsistency and ambiguity arising from a single source sentence \cite{tu2018learning}. Recently, there have been few attempts to model the semantic information across sentences. The basic ideas are to store a handful of previous source or target sentences with context vectors \cite{jean2017does,wang2017exploiting} or memory components \cite{maruf2017document,tu2018learning}. However, these methods have several limitations. First, the very short view of the previous sentences (usually one or two sentence(s)) is not sufficient enough to catch long term dependencies across paragraphs and storing detailed translation history is computationally expensive. Second, in the real-world scenarios, input data of MT application is often isolated sentences, such as \emph{Google Translate}, where no cross-sentence contexts are provided. Moreover, translations generated by such document-level NMT models are not stable, effected by the sentences surrounding the current one to translate.
	
	To address these limitations, we model the semantic information across sentences by mimicking the human translation process. In real scenarios, there will always be sentences or fragments that the translator can understand the meaning but cannot write down the translations directly. The obstacle could be unfamiliar collocation, descriptions in specific language habits and slang. The usual solutions for human are: (1) paraphrasing the sentence in another way, with simpler and more colloquial terms in the source language, and (2) directly referring to the standard translations of the intricate sentence fragments. For example in Table \ref{table-example1}, the Chinese word "zaiyu" is not a common expression. A reference can either provide simple Chinese terms such as "daizhe rongyu" or directly offer the corresponding English translation "with honor". Therefore, if a good quality \emph{reference book} which covers various translation scenes is provided, it can definitely improve the performance of human translators.
	
	\begin{table}
		\centering
		\begin{tabular}{l|p{5cm}}
			\hline		
			\bf{source} & canjia dongaohui de faguo yundongyuan \textbf{zaiyu} fanhui bali. \\
			\bf{translation} & French athletes participating in winter olympics returned to paris \textbf{with honors}. \\
			\hline	
		\end{tabular}
		\caption{An example of sentence fragment that is hard to translate.}
		\label{table-example1}
		
	\end{table}
	
	To be specific, the motivation of this work can be summarized as two aspects corresponding to the two kinds of human reference processes. First, we aim to provide the machine translator with a reference during decoding, which contains all possible source sentence fragments that are semantically similar to the current one. If the system finds it hard to translate the source fragment, it can turn to translate the fragments in the reference. Second, we intend to offer the oracle translations of the current sentence fragments to translate.
	
\begin{figure*}
		\centering
		\includegraphics[width=14cm]{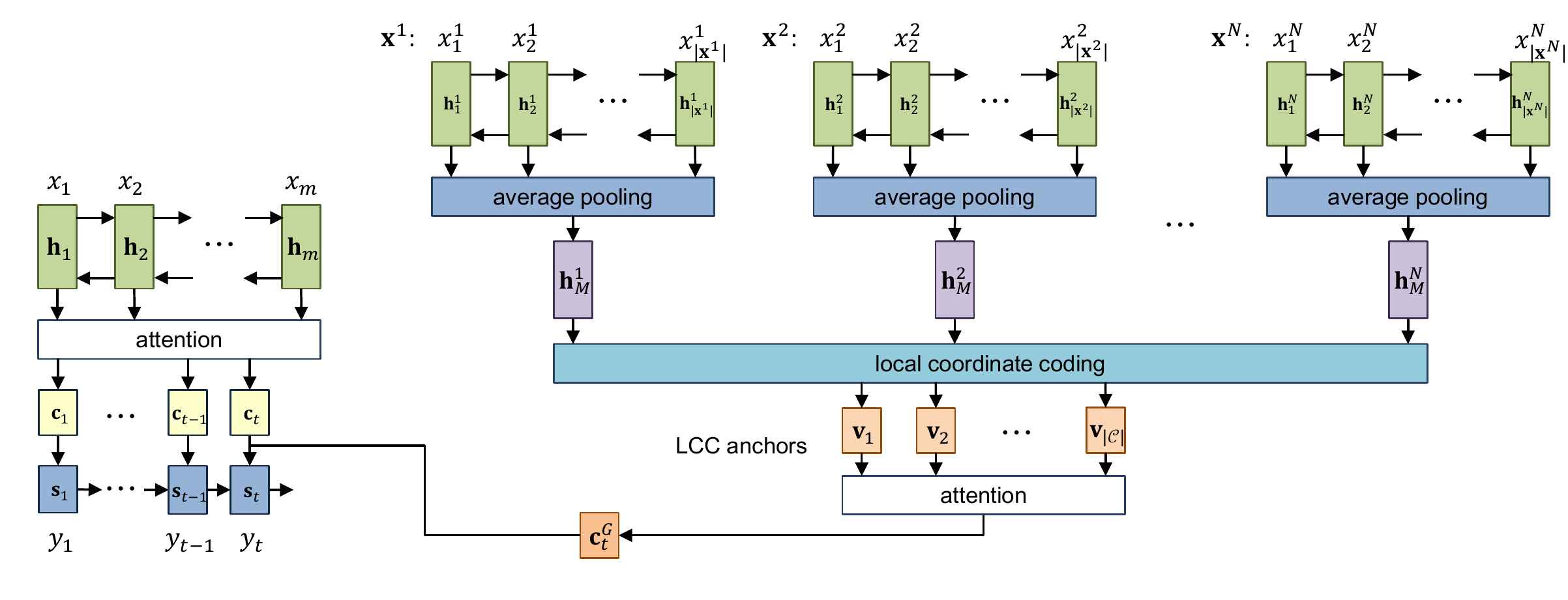}
		\caption{Framework of NMT with M-RefNet. $\mathbf{x}^{i}$ represents the $i$-th source sentence in the training corpus and $|\mathbf{x}^{i}|$ is the length of the sentence. The global context vector $\mathbf{c}_{t}^{G}$ can be regarded as a paraphrase of the current source context $\mathbf{c}_{t}$.} 
		\label{figure_arc_m}
	\end{figure*}

	In this paper, we propose a novel model namely Reference Network that incorporates the referring process into translation decoding of NMT. Instead of storing the detailed sentences or translation history, we propose to generate representations containing global monolingual and bilingual contextual information with Local Coordinate Coding (LCC) \cite{LLC}. Specifically, for solution (1), the hidden states of NMT encoder are coded by a linear combination of a set of anchor points in an unsupervised manner. The anchors are capable to cover the entire latent space of the source language seamlessly. For solution (2), we employ local codings to approximate the mapping from source and target contexts to the current target word with a supervised regression function. The local coding is then fed to the decoder to modify the update of the decoder hidden state. In this way, the translation decoding can be improved by offering the representation of a common paraphrase (Figure \ref{figure_arc_m}) or golden target translation (Figure \ref{figure_arc_b}). 
	
	We conduct experiments on NIST Chinese-English (Zh-En) and WMT German-Chinese (En-De) translation tasks. The experimental results indicate that the proposed method can effectively exploit the global information and improve the translation quality. The two proposed models significantly outperform the strong NMT baselines by adding only 9.3\% and 19.6\% parameters respectively.
	
	\section{Background}
	
	\subsection{Neural Machine Translation}
	Our model is built on the RNN-based NMT \cite{bahdanau2015neural}. However, since recurrent architecture is not necessary for our approach, the idea can also be applied to ConvS2S \cite{ConvS2S} and Transformer \cite{Trans}. We leave it for future work. Formally, let $\mathbf{x}=(x_{1},...,x_{m})$ be a given source sentence and $\mathbf{y}=(y_{1},...,y_{T})$ be the corresponding target sentence. NMT generates the target words sequentially by maximizing the probability of translation conditioned on the source sentence:
	\begin{equation}
	\hat{\mathbf{y}} = \mathop{\arg \max}_{\mathbf{y}}\sum_{t=1}^{T}\log p(y_t|\mathbf{x},\mathbf{y}_{<t} ).
	\end{equation}
	At each timestep, the generation probability is computed as
	\begin{equation}
	\label{eqn_out}
	p(y_t|\mathbf{x},\mathbf{y}_{<t} )={\rm softmax}(g (\mathbf{e}(y_{t-1}), \mathbf{s}_{t}, \mathbf{c}_{t})),
	\end{equation}
	where $g$ is a transformation function that outputs a vocabulary-sized vector, $\mathbf{e}(y_{t-1})$ is the embedding of previous target word $y_{t-1}$, $\mathbf{c}_{t}$ is the source context vector obtained by attention mechanism, and $\mathbf{s}_{t}$ is the $t$-th hidden state of NMT decoder, computed as:
	\begin{equation}
	\label{eqn_dec}
	\mathbf{s}_t=f_{d} (\mathbf{e}(y_{t-1}), \mathbf{s}_{t-1}, \mathbf{c}_{t}),
	\end{equation}
	where $f_{d}$ is a nonlinear activation. The source context $\mathbf{c}_{t}$ is typically a weighted sum of encoder hidden states as:
	\begin{equation}
	\mathbf{c}_{t}=\sum_{i=1}^{m}\alpha_{ti}\cdot\mathbf{h}_{i},
	\end{equation}
	where attention score $\alpha_{ti}$ is the alignment vector of the $i$-th source word $x_i$ and the $t$-th target word $y_t$:
	\begin{equation}
	\alpha_{ti}={\rm softmax}(\mathbf{v}_{\alpha}^{\top} {\rm tanh}(\mathbf{W}_{\alpha} \mathbf{s}_{t-1}+\mathbf{U}_{\alpha}\mathbf{h}_{i})).
	\end{equation}
	where $\mathbf{W}_\alpha$, $\mathbf{U}_\alpha$ and $\mathbf{v}_{\alpha}$ are trainable matrices or vectors. $\mathbf{h}_{i}$ is the annotation of $x_{i}$ computed by the NMT encoder. The encoder, generally implemented as a bi-directional RNN, encodes the input sentence into a sequence of source hidden states $\mathbf{h}=(\mathbf{h}_{1},...,\mathbf{h}_{m})$ where $\mathbf{h}_{i}$ is obtained by concatenating the forward hidden state $\overrightarrow{\mathbf{h}_{i}}$ and backward one $\overleftarrow{\mathbf{h}_{i}}$ at timestep $i$.
	
	According to the above formulations, conventional NMT models translate sentences independently. However, human translators usually tend to seek for reference materials when in trouble. Motivated by such common human behaviors, we propose Reference Network to provide global information as a \emph{reference book} in two ways. First, the model utilizes all source hidden states to paraphrase current source sentence. Second, the model directly provides the target word $\tilde{y_{t}}$ according to the rest translation samples in the training corpus. Since it is impossible to store all information directly, we leverage local coordinate coding (LCC) to compress the semantics into a latent manifold. 
	
	\subsection{Local Coordinate Coding}
	With the assumption that data usually lies on the lower dimensional manifold of the input space, the manifold approximation of high dimensional input $\mathbf{x}$ can be defined as a linear combination of surrounding anchor points as:
	\begin{equation}
	\mathbf{x}\approx\gamma(\mathbf{x})=\sum_{\mathbf{v}}\gamma_{\mathbf{v}}(\mathbf{x})\mathbf{v},
	\label{equation_lcc}
	\end{equation}
	where $\mathbf{v}$ is an anchor point and $\gamma_{\mathbf{v}}$ is the weight corresponding to $\mathbf{v}$ such that
	
	\begin{equation}
	\sum_{\mathbf{v}}\gamma_{\mathbf{v}}(\mathbf{x})=1.
	\end{equation}
	According to the definitions, it is proved in \cite{LLC} that if the anchor points are localized enough, any $(l_{\alpha}, l_{\beta})$-Lipschitz smooth function $f(\mathbf{x})$ defined on a lower dimensional manifold $\mathcal{M}$ can be globally approximated by a linear combination of the function values of a set of the anchors $\mathcal{C}$ as:
	
	\begin{equation}
	f(\mathbf{x})\approx\sum_{\mathbf{v}\in\mathcal{C}}\gamma_{\mathbf{v}}(\mathbf{x})f(\mathbf{v}),
	\end{equation}
	with the upper bound of the approximation error:
	\begin{equation}
	\begin{split}
	l_{\alpha}&\lVert\mathbf{x}-\sum_{\mathbf{v}\in\mathcal{C}}\gamma_{\mathbf{v}}(\mathbf{x})\mathbf{v}\rVert\\ +& \sum_{\mathbf{v}\in\mathcal{C}}l_{\beta}|\gamma_{\mathbf{v}}(\mathbf{x})|\lVert \mathbf{v}-\sum_{\mathbf{v}\in\mathcal{C}}\gamma_{\mathbf{v}}(\mathbf{x})\mathbf{v} \rVert^{2}.
	\end{split}
	\end{equation}
	
	\begin{figure*}
		\centering
		\includegraphics[width=14cm]{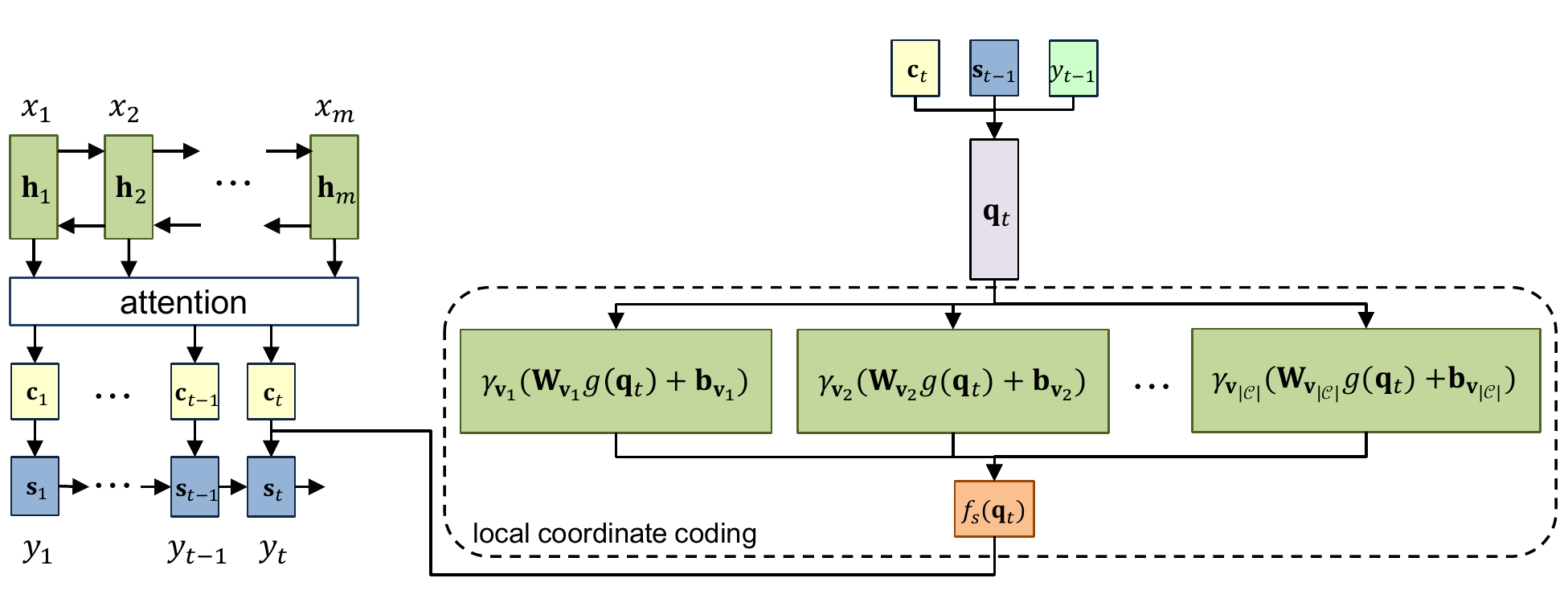}
		\caption{Framework of NMT with B-RefNet. The output $f_s(\mathbf{q}_{t})$ of RefNet can be regarded as an approximation of current target word embedding $\mathbf{e}(y_{t})$.}
		\label{figure_arc_b}
	\end{figure*}
	
	\section{Reference Network}
	
	In this section, we present our proposed Reference Network (RefNet).
	
	\subsection{Overview}
	
	We propose two models which explore the global information from the training data in different manners as illustrated by Figure \ref{figure_arc_m} and Figure \ref{figure_arc_b}. 
	
	The monolingual reference network (M-RefNet) provides a global source context vector to paraphrase the current context $\mathbf{c}_{t}$ based on all other source sentences. To be specific, we train several unsupervised anchors as the bases of the semantic space of source contexts and each source sentence in the training corpus can be represented by a weighted sum of the anchors.
	
	The bilingual reference network (B-RefNet) generates a referable target embedding according to all sentence pairs in the training corpus to guide output sequence generation.
	Concretely, we formulate the translation process as a mapping from source and target contexts ($\mathbf{c}_{t}$ and $\mathbf{s}_{t-1}$) to the current target word embedding $\mathbf{e}(y_{t})$. B-RefNet learns this mapping with a supervised regression function derived from LCC.
	
	It should be noted that the corpus from which the reference vectors ($\mathbf{c}_{t}^{G}$ or $f_s(\mathbf{q}_{t})$) are learned can be any monolingual or bilingual data, and the translations generated are relatively effected by the quality of the corpus. In this work, we constrain it as the training corpus for convenience and a fair comparison with the related work.
	
	\subsection{Monolingual Referent Network}
	
	In this section, we seek to improve NMT by rephrasing the source sentence. Instead of storing all source contexts, we regenerate the source contexts from a learned manifold with a combination of a fixed number of anchor points. Formally, given any source sequence $\mathbf{x}$ with length $m$ in the training samples, let $\mathbf{h}=(\mathbf{h}_{1},...,\mathbf{h}_{m})$ denotes the hidden states generated by the NMT encoder. We firstly obtain the representation of the source sentence $\mathbf{h}_{M}$ via a \emph{mean-pooling} operation. According to the definition of LCC, it can be assumed that $\mathbf{h}_{M}\approx\gamma(\mathbf{h}_{M})$ where $\gamma(\mathbf{h}_{M})$ is the local coordinate coding of $\mathbf{h}_{M}$, computed as:
	\begin{equation}
	\gamma(\mathbf{h}_{M}) = \sum_{j=1}^{|\mathcal{C}|}\gamma_{j}({\mathbf{h}_M})\mathbf{v}_{j}.
	\end{equation}
	Here, $\mathbf{v}_{j}$ is the $j$-th anchor point. The coefficient $\gamma_{j}(\mathbf{h}_{M})$ is used to measure the weight of anchor point $\mathbf{v}_{j}$ corresponding to $\gamma(\mathbf{h}_{M})$. In conventional manifold learning methods, $\gamma_{j}(\mathbf{h}_{M})$ is generally computed with distance measure. And to achieve localization, the coefficients corresponding to anchor points out of the neighbors of $\mathbf{h}_{M}$ are set to zero. However, it is hard to train in deep neural network using stochastic gradient methods. Inspired by the attention mechanism \cite{bahdanau2015neural}, we propose to employ an attention layer to obtain the weights:
	\begin{equation}
	\gamma_{j}(\mathbf{h}_{M})=\frac{\exp(s(\mathbf{h}_{M},\mathbf{v}_{j}))}{\sum_{j=1}^{|\mathcal{C}|}\exp(s(\mathbf{h}_{M},\mathbf{v}_{j}))},
	\end{equation}
	where $s(\cdot)$ is a score function.
	Here, we propose a tri-nonlinear score function which has been proven especially effective in the experiments:
	\begin{equation}
	\begin{split}
	\label{trilinear}
	s(\mathbf{h}_{M},\mathbf{v}_{j})&=\mathbf{v}_{s}^{\top} {\rm tanh}(\mathbf{W}_{s} \mathbf{v}_{j}+\mathbf{U}_{s}\mathbf{h}_{M}\\&+\mathbf{V}_{s}(\mathbf{v}_{j}\circ\mathbf{h}_{M})),
	\end{split}
	\end{equation}
	where $\mathbf{W}_{s}$, $\mathbf{U}_{s}$, $\mathbf{V}_{s}$ and $\mathbf{v}_{s}$ are trainable parameters. $\circ$ is the element-wise multiplication, and dimension of any anchor point should be the same to $\mathbf{h}_{M}$.
	
	To find the optimal anchor point, \emph{localization measure} \cite{LLC} is employed as the optimization object:
	\begin{equation}
	\label{local_measure}
	\begin{split}
	\mathop{\min}_{\gamma, \mathcal{C}}l_{\alpha}\left\|\mathbf{h}_{M}-\gamma(\mathbf{h}_{M})\right\|+\\l_{\beta}\sum_{j=1}^{|\mathcal{C}|}|\gamma_{j}(\mathbf{h}_{M})|\left\|\mathbf{v}_{j}-\gamma(\mathbf{h}_{M})\right\|^2.
	\end{split}
	\end{equation}
	
	Since any source sentence presentation $\mathbf{h}_{M}$ can be represented by the linear combination of the anchors, the trained anchor points can be regarded as the bases of the latent space of all source annotations, containing the global contextual information. Therefore, during translation decoding of NMT, we can drop the coefficient $\gamma$ and rephrase the source sentence only with the anchor points. Specifically, we apply an attention mechanism between current local contextual information and each anchor point $\mathbf{v}_{j}$ to get the global context as:
	\begin{equation}
	\mathbf{c}_{t}^{G} = \sum_{j=1}^{|\mathcal{C}|}\alpha_{tj}^{G}\mathbf{v}_{j},
	\end{equation}
	where $\alpha_{tj}^{G}$ is the attention score between current local contexts and the global context, computed as:
	\begin{equation}
	\label{att_m}
	\begin{split}
	\alpha_{tj}^{G}={\rm softmax}(\mathbf{v}_{\alpha}^{\top} {\rm tanh}(\mathbf{W}_{\alpha} \mathbf{s}_{t-1}\\+\mathbf{U}_{\alpha}\mathbf{c}_{t}+
	\mathbf{V}_{\alpha}\mathbf{v}_{j})).
	\end{split}
	\end{equation}
	Once the global context $\mathbf{c}_{t}^{G}$ is obtained, we feed it to decoder states:
	\begin{equation}
	\label{dec_m}
	\mathbf{s}_t=f_{d} (\mathbf{e}(y_{t-1}), \mathbf{s}_{t-1}, \mathbf{c}_{t}, \mathbf{c}_{t}^{G}),
	\end{equation}
	where $\mathbf{c}_{t}$ encodes the local contextual information and $\mathbf{c}_{t}^{G}$ contains the global monolingual information from all source sentences in the training corpus. When the model has trouble to translate some words or sentence fragments, it can refer to $\mathbf{c}_{t}^{G}$ to gain the richer source contextual information.
	
	\subsection{Bilingual Reference Network}
	
	The bilingual model is proposed to improve NMT by providing a golden translations according to rest samples in the training corpora. To be specific, once source context $\mathbf{c}_{t}$ and target context $\mathbf{s}_{t-1}$ are obtained, we hope to provide a referable prediction $\mathbf{e}(\tilde{y_t})$ of the current target word embedding $\mathbf{e}(y_t)$ according to other sentence pairs in the training data for the decoder.
	
	The functionality of the NMT decoder during translation (Eq.\ref{eqn_out} and Eq.\ref{eqn_dec}) is totally a function that maps the source context $\mathbf{c}_{t}$, target context $\mathbf{s}_{t-1}$ and last target word $y_{t-1}$ to current target $y_{t}$. NMT takes it as a classification problem, using $\rm{tanh}$ or other gated RNN unit to implement this function. In this work, we propose a much stronger model in information expression, that regrades the problem as regression:
	\begin{gather}
	\mathbf{q}_{t}=[\mathbf{e}(y_{t-1})^{\top},\mathbf{s}_{t-1}^{\top},\mathbf{c}_{t}^{\top}]^{\top},\\
	\label{reg_anc}
	\mathbf{e}(y_{t})\approx f_{s}(\mathbf{q}_{t})=\mathbf{W}(\mathbf{q}_{t})g(\mathbf{q}_{t})+\mathbf{b}(\mathbf{q}_{t}),
	\end{gather}
	where $g$ is a transformation function that transforms $\mathbf{q}_{t}$ to a anchor-size vector, $\mathbf{W}$ and $\mathbf{b}$ are the weight matrix and bias vector of the regression function. The weight and bias are allowed to vary according to the input $\mathbf{q}_{t}$, which makes the function capable of mapping each $\mathbf{q}_{t}$ to the corresponding $\mathbf{e}(y_{t})$ precisely. However, it is impossible to store the weight and bias for every $\mathbf{q}_{t}$ computed within the training data. Therefore, we approximate the weight and bias function in Eq.\ref{reg_anc} using local coordinate coding as:
	\begin{gather}
	\label{output_pre}
	f_s(\mathbf{q}_{t})=\sum_{j=1}^{|\mathcal{C}|}\gamma_{j}(\mathbf{q}_{t})\left(\mathbf{W}_{\mathbf{v}_{j}}g(\mathbf{q}_{t}) + \mathbf{b}_{\mathbf{v}_{j}}\right),
	\end{gather}
	where $\mathbf{v}_{j}\in{\mathcal{C}}$ is an anchor point, $\mathbf{W}_{\mathbf{v}_{j}}$ and $\mathbf{b}_{\mathbf{v}_{j}}$ are trainable parameters corresponding to $\mathbf{v}_{j}$, and $\gamma_{j}(\mathbf{q}_{t})$ is the weight function, computed as:
	\begin{equation}
	\gamma_{j}(\mathbf{q}_{t})=\frac{\exp(s(\mathbf{q}_{t},\mathbf{v}_{j}))}{\sum_{j=1}^{|\mathcal{C}|}\exp(s(\mathbf{q}_{t},\mathbf{v}_{j}))}.
	\end{equation}
	Similar to M-RefNet, the score is computed by the tri-nonlinear function as:
	\begin{equation}
	\label{tri-nonlinear}
	s(\mathbf{q}_{t},\mathbf{v}_{j})=\mathbf{v}_{b}^{\top} {\rm tanh}(\mathbf{W}_{b} \mathbf{v}_{j}+\mathbf{U}_{b}\mathbf{q}_{t}+\mathbf{V}_{b}(\mathbf{v}_{j}\circ\mathbf{q}_{t})).
	\end{equation}
	Here, $f_s(\mathbf{q}_{t})$ can be regarded as an approximation of $\mathbf{e}(y_{t})$ based on all the sentence pairs in the training data. Therefore, we feed the function value to the decoder state to guide sentence generation:
	\begin{equation}
	\mathbf{s}_t=f_{d} (\mathbf{e}(y_{t-1}), \mathbf{s}_{t-1}, \mathbf{c}_{t}, f_{s}(\mathbf{q}_{t})).
	\end{equation}
	
	The optimal weight matrices and anchor points are obtained by minimizing the hinge loss for each sentence pair $(\mathbf{x},\mathbf{y})$ as:
	\begin{equation}
	\mathcal{L}_{M}=\sum_{t=1}^{|\mathbf{y}|}\left\|\mathbf{e}(y_{t})-f_{s}(\mathbf{q}_{t})\right\|^{2}+\lambda_{M}\sum_{j=1}^{|\mathcal{C}|}\left\|\mathbf{W}(\mathbf{v}_{j})\right\|^2.
	\end{equation}
	
	\subsection{Training and Inference}
	Stage-wise training strategies have been proven to be efficient when system is relative complicated by plenty of recent work \cite{maruf2017document,tu2018learning}. In this work, we first pre-train a standard NMT on a set of training examples $\{[\mathbf{x}^{n},\mathbf{y}^{n}]\}_{n=1}^{N}$ as initialization for training the added parameters in our proposed models. 
	
	Let $\theta=\{\theta_{E},\theta_{D}\}$ denote the parameters of the standard NMT, where $\theta_{E}$ and $\theta_{D}$ are parameters of the standard encoder and decoder (including attention model) respectively. For M-RefNet, the stage following NMT training is to obtain the weight vectors $\gamma$ and anchor points $\mathcal{C}$ related to all training sentence representations $\mathbf{h}_{M}$ by minimizing \emph{localization measure} (Eq.\ref{local_measure}). Then we fix the trained anchor points and encoder, and only fine-tune the decoder $\theta_{D}$ and the added parameters $\theta_{M}$ related to the monolingual reference network (Eq.\ref{att_m} and Eq.\ref{dec_m}):
	\begin{equation}
	\mathop{\max}_{\theta_{D},\theta_{M}}\sum_{n=1}^{N}\left[\log P(\mathbf{y}^{n}|\mathbf{x}^{n};\theta,\theta_{M},\gamma)\right].
	\end{equation}
	
	To train B-RefNet efficiently, we fix the trained parameters of the standard NMT and only update the added parameters $\theta_{B}$ including all weight matrices and biases related to local coordinate coding (Eq.\ref{output_pre} and Eq.\ref{tri-nonlinear}). The training object is:
	\begin{equation}
	\label{loss_b}
	\mathop{\max}_{\theta_{B}}\sum_{n=1}^{N}\left[\log P(\mathbf{y}^{n}|\mathbf{x}^{n};\theta,\theta_{B})-\lambda\mathcal{L}_{M}\right],
	\end{equation}
	where $\lambda$ is a hyper-parameter that balances the preference between likelihood and hinge loss.
	
	During inference, all parameters related to LCC are fixed. Therefore, the work can be regarded as a \emph{static} approach, compared with the conventional document-level NMT. That means, the final translation is only effected by the reference corpus but not by the sentences surrounding the current one to translate. Naturally, there leaves a question that how it influences the quality of translations when various reference corpus is chosen. We leave it in future work and only use the training corpus in this paper.
	
	\section{Experiments}
	We evaluate the reference network models on two translation tasks, NIST Chinese-English translation (Zh-En) and WMT English-German translation (En-De).

	\begin{table}
		
		\centering
		\begin{tabular}{l|cccc}
			
			System &    MT05 &       MT06 &       MT08 &        Avg \\
			\hline
			\hline
			Dl4mt & 32.88 & 32.30 & 25.97 & 30.38 \\
			\hline
			NMT & 35.76 & 34.82 & 27.86 & 32.81 \\
			\hline
			CS-NMT & 36.63 & 36.41 & 29.47 & 34.17 \\
			
			LS-NMT & 36.46 & 36.99 & 29.73 & 34.39 \\
			
			CC-NMT & 36.65 & 37.08 & 29.71 & 34.48 \\
			
			DC-NMT & 36.82 & 36.73 & 29.83 & 34.46 \\
			
			\hline
			\multicolumn{5}{c}{\emph{This work}} \\
			\hline

			M-RefNet & 37.31 & 37.72 & 30.41 & 35.15\\
			
			B-RefNet & \textbf{37.71} & \textbf{37.99} & \textbf{30.80} & \textbf{35.50} \\
			
		\end{tabular}
		\caption{BLEU scores of different models on Zh-En.}  
		\label{table-zh-en}
	\end{table}
	
	\subsection{Settings}
	\paragraph{Datasets}
	For Zh-En, we choose 1.25M sentence pairs from LDC dataset\footnote{The corpus contains LDC2002E18, LDC2003E07, LDC2003E14, Hansards portion of LDC2004T07, LDC2004T08 and LDC2005T06} with 34.5 English words and 27.9M Chinese words. NIST MT02 is chosen as the development set, and NIST MT05/06/08 as test sets. Sentences with more than 50 words are filtered and vocabulary size is limited as 30$k$. We use case-insensitive BLEU score to evaluate Zh-En translation performance.
	For En-De, the training set is from \cite{Luong2015Effe} which contains 4.5M bilingual pairs with 116M English words and 100M German words. BPE \cite{sennrich2016neural} is employed to split the sentence pairs into subwords and we limit the vocabulary as 40$k$ sub-words units. Newstest2012/2013 are chosen for developing and Newsetest2014 for test.
	case-sensitive BLEU\footnote{https://github.com/moses-smt/mosesdecoder /blob/master/scripts/generic/multi-bleu.perl} is employed as the evaluation metric.
	
	\paragraph{Models}
	We evaluate our RefNet with different structures on Zh-En and En-De. For Zh-En we choose the typical attention-based recurrent NMT model \cite{bahdanau2015neural} as initialization, which consists of a bi-directional RNN-based encoder and a one layer RNN decoder. The dimensions of embedding and hidden state are 620 and 1000 respectively. For En-De, deep linear associative unit model (DeepLAU) \cite{wang2017lau} is chosen as the base model. Both the encoder and decoder consist of 4-layer LAUs. All embedding and hidden states are 512-dimensional vectors. Moreover, we use layer normalization \cite{ba2016layer} on all layers. For both architectures, the number of anchor points is 100 for M-RefNet and 30 for B-RefNet. The anchor dimension of B-RefNet is set to 100. The hyper-parameter $\lambda$ in Eq.\ref{loss_b} is set to 1. The norm of gradient is clipped to be within $[-1,1]$ and dropout is applied to embedding and output layer with rate 0.2 and 0.3 respectively. When generating translations, we utilize beam search with beam size 10 on Zh-En and 8 on En-De.
	
	\subsection{Results on Chinese-English Translation}
	The standard attention-based NMT model is chosen as the baseline and initialization of our models. Moreover, we also list the results of the open-source Dl4mt and re-implementations of the following related work for comparison:
	\begin{itemize}
		\item Cross-sentence context-aware NMT (CS-NMT) \cite{wang2017exploiting}: A cross-sentence NMT model that incorporates the historical representation of three previous sentences into decoder.
		\item LC-NMT \cite{jean2017does}: A NMT model that concurrently encodes the previous and current source sentences as context, added to decoder states.
		\item NMT augmented with a continuous cache (CC-NMT) \cite{tu2018learning}: A NMT model armed with a cache\footnote{Cache size is set to 25.} which stores the recent translation history.
		\item Document Context NMT with Memory Networks (DC-NMT) \cite{maruf2017document}: A document-level NMT model that stores all source and target sentence representations of a document to guide translation generating\footnote{LDC training corpora contains nature boundaries. However document range is not clear for NIST test data. We use clustering and regard each class as a document. Dimension of document context is set to 1024.}.
	\end{itemize}
	All the re-implemented systems share the same settings with ours for fair comparisons.
	
	\subsubsection{Main Results}
	Results on Zh-En are shown in Table \ref{table-zh-en}. The baseline NMT significantly outperforms the open-source Dl4mt by 2.43 BLEU points, indicating the baseline is strong. Our proposed M-RefNet and B-RefNet improve the baseline NMT by 2.34 and 2.69 BLEU respectively and up to 2.90 and 3.17 BLEU on NIST MT06, which confirms the effectiveness of our proposed reference networks. Overall, B-RefNet achieves the best performance over all test sets

	\begin{table}
		
		\centering
		\begin{tabular}{c|lc|c|c}
			\multirow{2}{*}{\#} & \multirow{2}{*}{System} & \multirow{2}{*}{\#Para} & \multicolumn{2}{c}{Speed } \\
			\cline{4-5}
			& & & Train & Test \\
			\hline
			\hline
			0 & NMT & 71.1M & 3590.4 & 114.21 \\
			\hline
			1 & CS-NMT & 95.7M & 747.5 & 97.10 \\
			2 & LC-NMT & 96.8M & 1983.5 & 70.11 \\
			3 & CC-NMT & 75.1M & \textbf{2844.7} & 113.09 \\
			4 & DC-NMT & 86.2M & 2093.6 & 54.07 \\
			\hline
			5 & M-RefNet & 77.7M & 2563.98 & \textbf{113.26} \\
			6 & B-RefNet & 85.1M & 2191.4 & 104.07 \\

		\end{tabular}
		\caption{Statistics of parameters, training speed (sentences/minute) and testing speed (words/second).}
		\label{table-para}
		
	\end{table}
	
	\begin{figure}
		\centering
		\includegraphics[width=7.8cm]{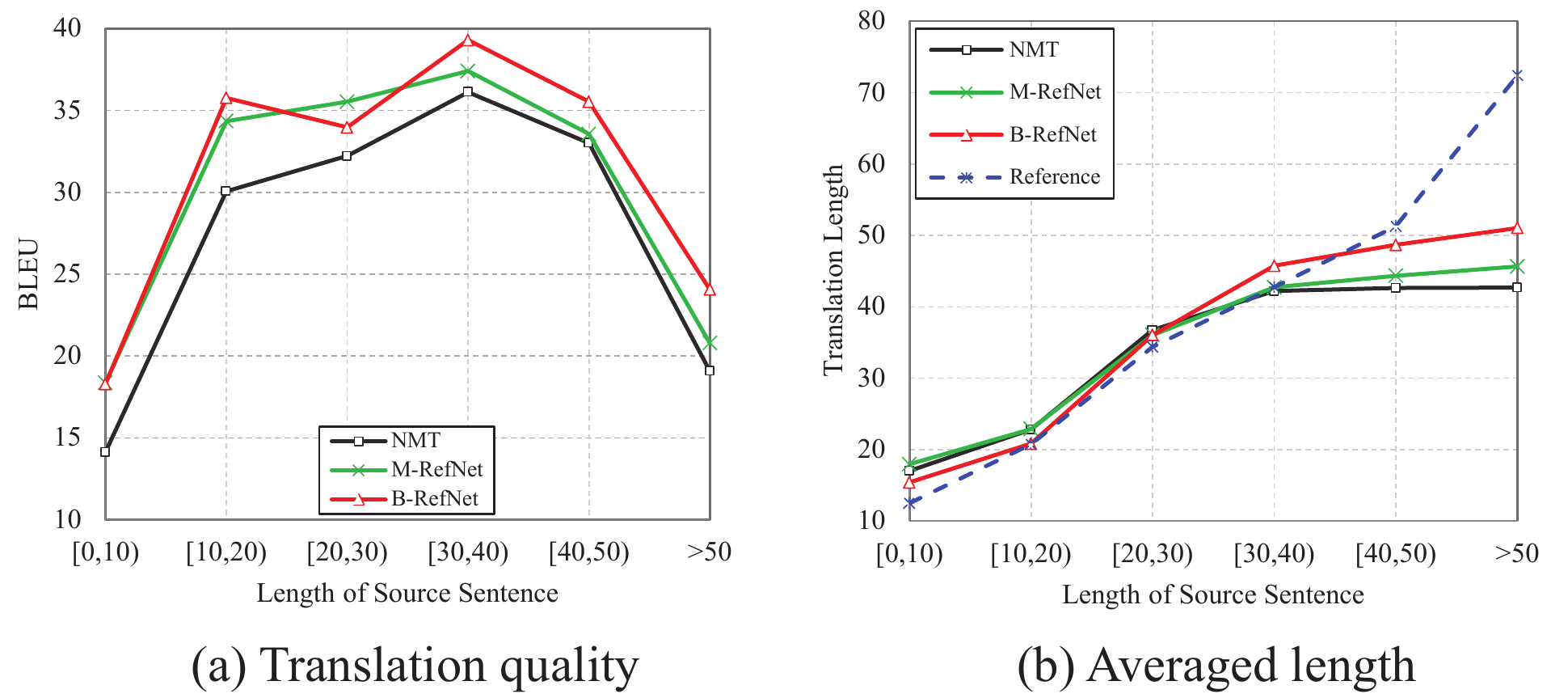}
		\caption{Translation quality and averaged length of the translations as source sentences become longer.}
		\label{figure_length}
	\end{figure}
	
	Compared with the related work which incorporate document-level information NMT, our proposed models still have a significant advantage. Compared to the best performance achieved by the related work (CC-NMT), M-RefNet and B-RefNet outperform it over all test sets and 
	gain improvements of 0.77 BLEU and 1.02 BLEU in average. The possible reason is that all the related work only leverage a small range of the document-level information, limited by model complexity and time consuming. In contrast, our models are capable to express all information with more abstract representations. According to the results, though the information is deeply compressed in our models, it is still effective.
	
	\subsubsection{Analysis}
	
	\begin{table*}
		\centering
		\begin{tabular}{l||p{13.5cm}}
			\hline
			Source & agenting zantong yige zhongguo de \textbf{lichang} . \\
			\hline
			Reference & argentina supports the " one china " \textbf{policy}. \\
			\hline
			NMT & argentina agrees with china ' s \emph{stand on} the one china . \\
			\hline
			M-RefNet & the argentine government supports the one china \textbf{position} .\\
			\hline
			B-RefNet & argentina supports the one china \textbf{policy} .  \\
			\hline
			\hline
			Source & yindu bianfang minbing 2 yue 17 ri , \textbf{jiaqiang le dui niboer bianjie de xunluo} . \\
			\hline
			Reference & on february 17 , the indian border security force \textbf{stepped up patrols along the border with nepal} . \\
			\hline
			NMT & on 17 february , indian border defense militia [ \emph{UNK UNK UNK UNK UNK} ] . \\
			\hline
			M-RefNet & the indian border defense militia , on 17 february , \textbf{strengthened the patrol of nepal ' s border} .\\
			\hline
			B-RefNet & the indian border defense militia \textbf{has stepped up patrols on the nepalese border} on 17 february . \\
			\hline
		\end{tabular}
		\caption{Comparison on translation examples. The translation errors are highlighted with italic and the correct ones are highlighted with bold type.}
		\label{table-case}
	\end{table*}
	
	\paragraph*{Parameters and Speed}
	The number of parameters and speed of each model are listed in Table \ref{table-para}. It can be seen that M-RefNet only introduces 6.6M additional parameters while B-RefNet introduces relative larger number of parameters (14M). Considering training process, both M-RefNet and B-RefNet are quite efficient and the training speeds are little slower than CC-NMT, for the added amount of parameters is quite small compared to the baseline NMT and related systems. In terms of decoding, both proposed models do not slow down the translation speed obviously and M-RefNet achieves the fastest speed over all systems except the baseline NMT. The reason is that our models do not incorporate additional previous sentences or interact with extra memory as the relevant document-level systems. Furthermore, though the training speed and number of parameters of B-RefNet and DC-NMT are similar, B-RefNet gains a twice faster translation speed, because that DC-NMT needs a two-pass translation process to fill the memory cells.
	
	\paragraph{Length Analysis}
	We follow \cite{Luong2015Effe} to group sentences with similar lengths and compute the BLEU score of each group, as shown in Figure \ref{figure_length}. The reason for the falling of BLEU in the last group ($>$50) is that sentences longer than 50 are removed during training. From this figure, we can see that our proposed models outperform the baseline NMT in all ranges of length. Moreover, translations generated by M-RefNet and B-RefNet have more similar lengths to the references compared with the baseline NMT.
	
	\paragraph*{Case Study}
	Table \ref{table-case} shows the translation examples on Zh-En. In the first case, the Chinese word "lichang" (standpoint, position, or policy) is incorrectly interpreted as "stand on" by NMT. Both M-RefNet and B-RefNet generate legible translations while translation from B-RefNet is more precise. This is because the word pair ("lichang", "policy") appear somewhere in the training data and is leveraged by the systems according to the contexts. This phenomenon is similar in the second case. Translation given by NMT is not readable. In contrast, M-RefNet generates the core verb "strengthened" and B-RefNet provides a more accurate collocation "stepped up patrols".
	
	\begin{table*}
		\centering
		\begin{tabular}{l|ll|c} 
			& System & Architecture &BLEU\\
			\hline
			\hline
			0 & GNMT & 8-layer LSTM encoder and decoder& 24.60\\
			1 & Robust NMT & 2-layer GRU encoder and decoder + adversarial training & 25.26 \\
			2 & ConvS2S &15-layer CNN encoder and decoder & 25.16\\
			
			3 & Transformer (big) & 6-layer encoder and decoder + 16-head self-attention & 28.40\\
			\hline
			\multicolumn{4}{c}{\emph{This work}}\\
			\hline
			4 & DeepLAU & & 24.37\\
			5 & M-RefNet &4-layer LAU encoder and decoder& 25.66\\
			6 & B-RefNet && \textbf{26.16} \\
			
		\end{tabular}
		\caption{ Translation quality on En-De.}	
		\label{table-en-de}
	\end{table*}

	\subsection{Results on English-German Translation}
	On this task, DeepLAU \cite{wang2017lau} is chosen as the baseline and also used as the pre-trained model. We list the translation performance of our models and some existing NMT systems in Table \ref{table-en-de}. All the systems except for Robust NMT \cite{cheng2018robust} have a deep architecture with no less than 4 layers while Robust NMT introduces a additional discriminator for adversarial training. From the table, we can observe that our strong baseline DeepLAU is comparable to Google's neural machine translation (GNMT) \cite{GNMT}. M-RefNet outperforms the baseline by 1.29 BLEU points and B-RefNet achieves slightly better performance with a 1.79 BLEU improvement, which is consistent to the results on Zh-En. Compared with the SOTA deep NMT systems, both M-RefNet and B-RefNet outperform GNMT and even obtain comparable performance with ConvS2S \cite{ConvS2S} and Transformer \cite{Trans} which have much deeper architectures with relative much more parameters. Since the reference networks do not rely on the recurrent structure, one interesting future direction is to apply our methods to such complicated models to bring further improvements.

	\section{Related Work}
	
	\paragraph*{Document-level Neural Machine Translation}
	There are few works that consider the document-level contextual information to improve typical NMT. \citet{jean2017does} propose to use a additional encoder to generate the latent representation of previous sentence as extra context for decoder and attention mechanism is also applied between the decoder state and previous context to get access to word-level information of the previous sentence. Contemporaneously, \citet{wang2017exploiting} extend NMT by adding two encoders to encode the previous sentences in word-level and sentence-level respectively. The last hidden state of encoders are considered as the summarization of a previous sentence and the group. \citet{bawden2017evaluating} employ multiple encoder s to summarize the antecedent and propose to combine the contexts with a gated function. However, these incorporated extra encoders bring large amount of parameters and slow down the translation speed. \citet{tu2018learning} propose to modify the NMT with light-weight key-value memory to store the translation history. However, due to the limitation of the memory size, the very short view on the previous (25 timesteps) is not sufficient to model the document-level contextual information. Additionally, \citet{maruf2017document} propose to capture the global source and target context of a entire document with memory network \cite{Graves2014Neural,wang2017memory}. Nevertheless, since the number of sentence pairs in a document could be enormous, storing all sentence with memory components could be very time and space consuming. More recently, \citet{miculicich2018document} and \citet{zhang2018improving} propose to improve Transformer by encoding previous sentences with extra encoders. The \emph{reference book} in this work can be regarded as a special kind of document context. However, there are two major differences between our approach and the above work. First, we encode the entire corpus into a handful of anchor points which is much more light-weight but concentrated to capture the global contextual information
	. Second, the global contexts in this work is \emph{static}. That means, given a sentence to translate, the final translation result only depends on the reference corpus, but not the sentences surrounding the current one.
	\paragraph*{Local Coding}
	There are a number of works on manifold learning \cite{roweis2000nonlinear,van2008kernel,LLC,ladicky2011locally}. The manifold learning methods approximate any point on the latent manifold with a linear combination of a set of localized anchor points relying on the assumption that high dimensional input usually lies on the lower dimensional manifold. \citet{agustsson2017anchored} utilize local coding into deep neural networks on age prediction from images and \citet{cao2018adversarial} exploit LCC for GAN \cite{goodfellow2014generative} to capture the local information of data. All these works focus on application of Computer Vision while we apply LCC in a Nature Language Processing task. To our knowledge, this is the first attempt to incorporate local coding into NMT modeling.
	
	\section{Conclusion and Future Work}
	In this work, we propose two models to improve the translation quality of NMT inspired by the common human behaviors, paraphrasing and consulting. The monolingual model simulates the paraphrasing process by utilizing the global source information while the bilingual model provides a referable target word based on other sentence pairs in the training corpus. We conduct experiments on Chinese-English and English-German tasks, and the experimental results manifest the effectiveness and efficiency of our methods.
	
	In the future, we would like to investigate the feasibility of our methods on non-recurrent NMT models such as Transformer \cite{Trans}. Moreover, we are also interested in incorporating discourse-level relations into our models.
	
	\section*{Acknowledgments}
	We would like to thank the reviewers for their valuable comments and suggestions.
	
	%\bibliography{acl2018}
	\bibliography{GlobalContext}{}
	\bibliographystyle{acl_natbib}
	
\end{document}